\documentclass[letterpaper, 10 pt, conference]{ieeeconf}  

\IEEEoverridecommandlockouts                              
\usepackage{graphics} 
\usepackage{amsmath} 
\usepackage{amssymb}  
\usepackage{mathtools}

\usepackage{amsthm}
\usepackage{color}
\usepackage{bm}
\usepackage[ruled]{algorithm2e} 
\usepackage{multirow}
\usepackage{hyperref}
\usepackage[backend=biber,
            hyperref=true,
            url=false,
            isbn=false,
            doi=false,
            backref=false,
            style=ieee,
            citestyle=numeric-comp,
            sorting=nyt,
            block=none]{biblatex}

\usepackage[font=small,skip=4pt]{caption}

\renewcommand{\bibfont}{\small}
\theoremstyle{plain}

\renewcommand{\baselinestretch}{0.97}

\title{\LARGE \bf Real-Time Model Predictive Control for Industrial Manipulators \\ with Singularity-Tolerant Hierarchical Task Control}

\author{Jaemin Lee$^{1}$, Mingyo Seo$^{2}$, Andrew Bylard$^{3}$, Robert Sun$^{3}$, and Luis Sentis$^{2}$
\thanks{We thank Dexterity and the HCRL personnel. Jaemin Lee and Mingyo Seo were interns for Dexterity Inc. during the summer of 2022. Also, Luis Sentis was a consultant for Dexterity Inc. during the summer of 2022.}
\thanks{$^{1}$J. Lee is with the Department of Mechanical and Civil Engineering, California Institute of Technology, Pasadena, CA, USA, \tt\small jaemin87@caltech.edu}%
\thanks{$^{2}$M. Seo and L. Sentis are with The University of Texas at Austin, Austin, TX, USA, \tt\small \{mingyo, lsentis\}@utexas.edu }
\thanks{$^{3}$A. Bylard, and R. Sun are with Dexterity Inc., Redwood City, CA, USA, \tt\small \{ andrew.bylard,robert\}@dexterity.ai}%
}

\begin{document}

\maketitle
\thispagestyle{empty}
\pagestyle{empty}

\begin{abstract}
This paper proposes a real-time model predictive control (MPC) scheme to execute multiple tasks using robots over a finite-time horizon. In industrial robotic applications, we must carefully consider multiple constraints for avoiding joint position, velocity, and torque limits. In addition, singularity-free and smooth motions require executing tasks continuously and safely. Instead of formulating nonlinear MPC problems, we devise linear MPC problems using kinematic and dynamic models linearized along nominal trajectories produced by hierarchical controllers. These linear MPC problems are solvable via the use of Quadratic Programming; therefore, we significantly reduce the computation time of the proposed MPC framework so the resulting update frequency is higher than $1$ kHz. Our proposed MPC framework is more efficient in reducing task tracking errors than a baseline based on operational space control (OSC). We validate our approach in numerical simulations and in real experiments using an industrial manipulator. More specifically, we deploy our method in two practical scenarios for robotic logistics: 1) controlling a robot carrying heavy payloads while accounting for torque limits, and 2) controlling the end-effector while avoiding singularities.   
\end{abstract}

\section{Introduction}
Robotic systems have been broadly utilized in automated industrial applications such as logistics. In these environments, it is important to guarantee the success and safety of manipulation tasks, such as packing, singulating, palletizing, and depalletizing \cite{echelmeyer2008robotics}. To perform the above tasks, robot manipulators need to perform fast and safe operations for grabbing, manipulating, and tossing boxes, objects, or parcels. These often require careful specification and tracking of task-space motions and hierarchical task-space objectives, for example to prioritize position control while maintaining sensible payload orientations throughout a trajectory. Carrying heavy payloads is also critical for robotic manipulators in terms of both their mechanical design \cite{baccelliere2017development} and their controller implementation \cite{aghili2009impedance}. In this paper, we tackle the above industrial manipulation problems to improve the task motion tracking performance of robots while generating fast, safe, and smooth motions.  

Operational Space Control (OSC) and Task Space Control (TSC) have been employed to generate dynamically consistent torque commands to effectively and safely track task trajectories \cite{khatib1987unified, nakanishi2008operational, mistry2012operational}. For instance, OSC improves the tracking performance of heavy manipulators by considering feedforward terms based on  robot dynamics \cite{maeda2011improving}. In addition, robot manipulators can generate compliant behaviors for safe manipulation by using TSC with null-space projection matrices in the presence of humans \cite{sadeghian2013task}. Many advanced approaches based on OSC and TSC allow manipulators to safely move in the vicinity of singularities \cite{kang2019singularity,lee2020task} or consider torque input saturation in operational space \cite{murtaza2021real, braun2019operational}. However, OSC and TSC are based on single-step optimizations resulting on myopic motions (only optimal locally at each time step) and are susceptible to abrupt changes due to the effect of the task specifications or constraints.

\begin{figure}
\centering
\includegraphics[width=\linewidth]{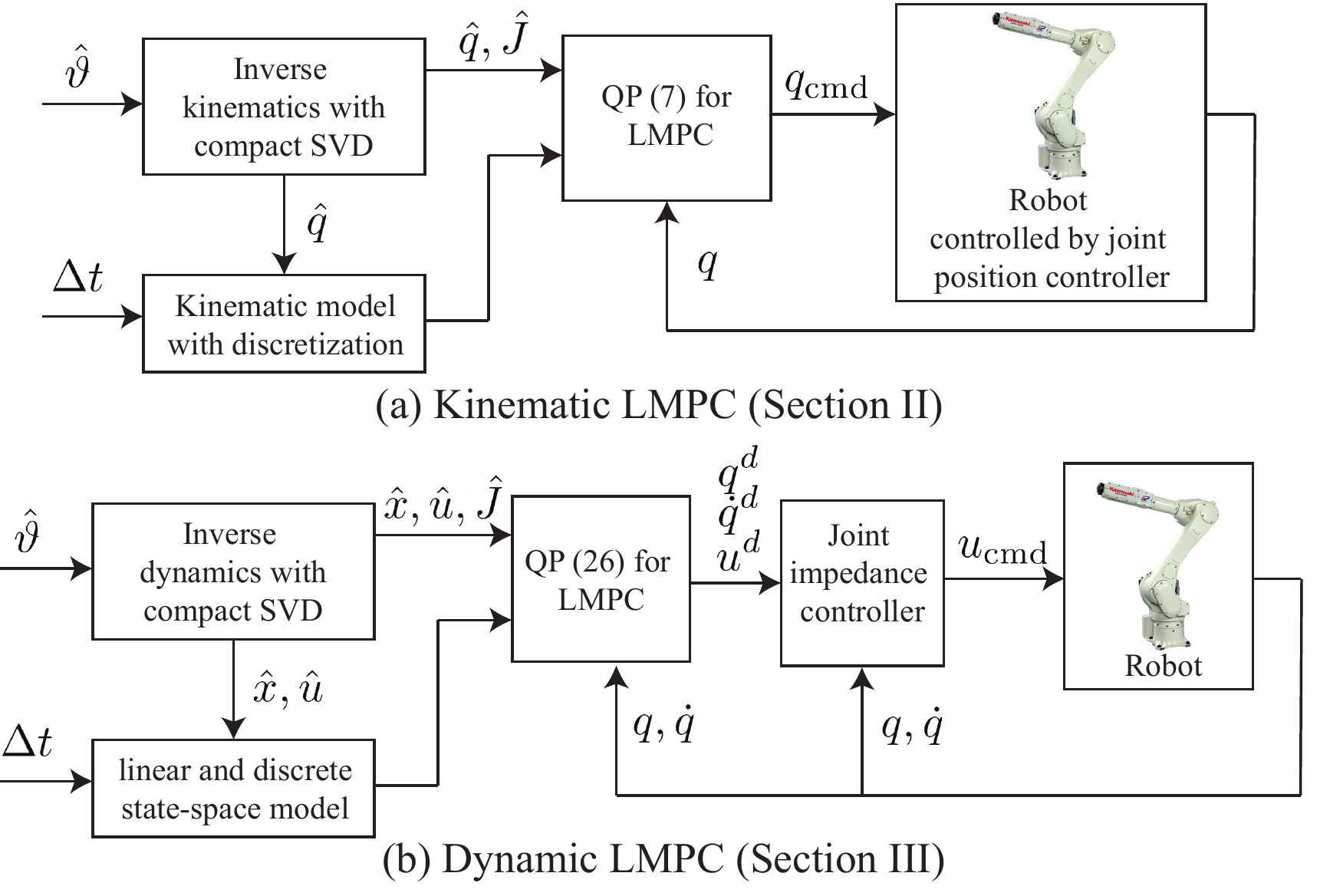}
\caption{{\bf Block diagrams of the proposed MPC approaches}: (a) Kinematic MPC for robots controlled by a joint position controller, (b) Dynamic MPC for torque-controlled robots. \vspace{-10pt} }
\label{Fig1}
\end{figure}

Model predictive control (MPC) has been frequently combined with impedance control \cite{bednarczyk2020model} or inverse dynamics control \cite{incremona2017mpc,nicolis2020operational} when controlling manipulators. Usually, the feedback MPC is updated at lower frequency ($20$ Hz - $50$ Hz), compared with the feedback control frequency ($400$ Hz - $1$ kHz) \cite{grandia2019feedback, koenemann2015whole}. However, an ideal MPC should execute with direct sensor feedback at a high-frequency update rate ($500$ Hz -  $1$ kHz) \cite{kleff2021high}. However, the dynamic models and constraint requirements imposed by robot manipulators result in nonlinear problem formulations. So it is difficult to solve the problems at fast rates using MPC, particularly when coupled with hierarchical task specifications which are increasingly in demand in industrial applications. MPC problems have previously been formulated as quadratically-constrained quadratic programs \cite{lee2020mpc} and sequential optimization problems \cite{sathya2022simple}, then solved via convex optimization. However, the speed of these MPC approaches are still not sufficient for implementation within a real-time control loop to provide the needed industrial performance. For industrial applications, it is extremely important to implement the hierarchical task control while improving the task tracking performance via MPC within a real-time control loop. Recent studies have tried to treat the nonlinear models efficiently \cite{nubert2020safe}, or to learn the terminal cost \cite{kang2022tracking} using data-driven techniques such as neural networks increasing their task tracking performance.
However, they are complicated to implement, they produce limited reduction of computation time, and they are missing key details of their computational performance such as dependence on the receding horizon and their control computer specifications.

Three significant issues arise when employing MPC in industrial manipulation: (1) dealing with nonlinear dynamics and cost functions, (2) dealing with hierarchical task specifications, and (3) guaranteeing stability and robustness to singularities while tracking task trajectories. To resolve the above issues, we aim to formulate kinematic and dynamic MPC approaches which can be executed with low-level controllers at a high-frequency update rate, as shown in Fig. \ref{Fig1}. Our framework uses inverse kinematic and dynamic control formulations in the operational space to generate nominal trajectories. Since the stability of OSC while performing hierarchical tasks is verified in \cite{dietrich2018hierarchical}, it is assumed that using OSC provides stable nominal trajectories as inputs for the proposed MPC. In addition, we enforce terminal state constraints to stabilize our MPC \cite{mayne2000constrained}. Based on the input nominal trajectories, we devise a simple formulation for MPC to reduce task tracking errors with additive cost terms for generating smooth motions. Our kinematic and dynamic MPC approaches are formulated as Quadratic Programming (QP) problems, which can be solved very fast. 

The main contributions of our work are as follows. First, we propose a framework that integrates hierarchical control, MPC, and  low-level control. In particular, our proposed MPC framework, which is for executing multiple tasks with hierarchy, is updated at the same fast rate as the low-level control loop. We verify that the proposed MPC is sufficiently fast to be executed at a $1$ kHz closed-loop update rate through both experiments and simulations. We also analyze the effect of the receding horizon on computation time.
Second, we report that our MPC results follow the task priority imposed by the hierarchical controller while reducing the task tracking error. We showcase simulations showing that the proposed MPC framework reduces task tracking errors when performing manipulation tasks with a heavy payload under torque saturation.
Third, our proposed MPC framework helps to avoid the involuntary termination of robot movement when the joint velocities exceed their limit when passing through singularities. We experimentally show that the proposed MPC generates stable and smooth behaviors in the vicinity of the singularities. For validation, we apply the proposed MPC framework to two industrial Kawasaki manipulators: the RS007N and RS020N models.  


The remainder of this paper is organized as follows. We propose our kinematic MPC and explain its implementation in Section II. Section III presents the proposed dynamic MPC with linearization and discretization in detail. In Section IV, we deploy the proposed MPC approaches to demonstrate two practical scenarios using industrial manipulators. Numerical simulations and experiments show that the proposed MPC approach reduces tracking errors and generates safe smooth motions.

\section{Kinematic MPC with Constraints}
\subsection{Kinematic Model with Discretization}
We perform simple discretization of the joint velocity and acceleration with a time interval $\Delta t$ (usually identical to the time interval for updating a low-level controller) as follows:
\begin{equation} \label{eq_01}
    \dot{q}_{i} = \frac{q_{i} - q_{i-1}}{\Delta t}, \quad \ddot{q}_{i} = \frac{q_{i} - 2q_{i-1} + q_{i-2}}{\Delta t^2}.
\end{equation}
Now, let us consider a task function $g: \mathbb{R}^{n} \mapsto \mathbb{R}^{n_g}$ which is $C^{1}$ and $\vartheta_{k} = g(q)$ with 
\begin{equation}
    \Delta \vartheta  = J \Delta q,
\end{equation}
where $J \in \mathbb{R}^{n_g\times n}$. For instance, the functional output of this task can be the position and orientation of the end-effector so that $\vartheta \in \mathrm{SE}(3)$. Then, the low-level controller will drive the robot to track the desired joint position trajectory.

Given a finite-time horizon $\mathbf{T} = [t_i,\: t_{f}]$ where $t_i < t_f$, we vertically concatenate the joint positions/velocities/accelerations such that $\mathbf{q} = [q_{i}^{\top},\: \cdots, \: q_{f}^{\top}]^{\top}$, $\dot{\mathbf{q}} = [\dot{q}_{i}^{\top},\: \cdots, \: \dot{q}_{f}^{\top}]^{\top}$, and $\ddot{\mathbf{q}} = [\ddot{q}_{i}^{\top},\: \cdots, \: \ddot{q}_{f}^{\top}]^{\top}$. Then, the joint velocity and acceleration in \eqref{eq_01} are expressed in compact form as
\begin{equation}
\begin{split}
    \dot{\mathbf{q}} = \mathbf{S}_v \mathbf{q} + \mathbf{v}, \quad
    \ddot{\mathbf{q}} = \mathbf{S}_a \mathbf{q} + \mathbf{a},
\end{split}
\end{equation}
where $\mathbf{S}_v$ and $\mathbf{S}_a$ are constant matrices. Assuming that the initial velocity and acceleration are zero, we can obtain components of the vectors $\mathbf{v}$ and $\mathbf{a}$ as constant vectors. As a result, we express the joint velocity and acceleration in linear affine form in terms of the joint position.  

\subsection{MPC formulation}
We aim to formulate a generic optimization problem for our kinematic MPC approach. Compared to OSC, our kinematic MPC generates smoother and more robust control commands by tightly integrating them with a low-level joint position controller. This section assumes that the low-level joint position controller is well-tuned and stable within the desired motion bandwidth. 

Given a task trajectory $\{\hat{\vartheta}_{0}, \cdots, \hat{\vartheta}_{N}\}$, the optimal control problem is formulated with a prediction horizon $n_p$. The cost function consists of task tracking errors, joint damping, and joint acceleration terms. In addition, the joint position and velocity limits are considered in the optimization problem 
\begin{equation} \label{kin_opt_original}
\begin{split}
    \min_{\mathbf{q}} \quad & \mathcal{J} = \sum_{k=i}^{i+n_p} (\hat{\vartheta}_k - g(q_k))^{\top}Q_{e}(\hat{\vartheta}_k - g(q_k)) \\
    & \qquad + \dot{q}_{k}^{\top}Q_{d} \dot{q}_{k} + \ddot{q}_{k}^{\top}Q_a \ddot{q}_{k} ,\\
    \textrm{s.t.}\quad& \dot{q}_{\min} \leq \dot{q}_{k} \leq \dot{q}_{\max},\\
    & q_{\min} \leq q_{k} \leq q_{\max}, \quad \forall k \in \{ i, \cdots, i +n_p\},
\end{split}
\end{equation}
where $Q_e \in \mathbb{R}^{n_g\times n_g}$, $Q_d \in \mathbb{R}^{n\times n}$, and $Q_a \in \mathbb{R}^{n \times n}$ denote the weighting matrices for the tracking task error, joint damping, and joint acceleration terms, respectively. Since the task mapping $g$ is a nonlinear functional mapping, the first term in the cost function, $(\hat{\vartheta}_k - g(q_k))^{\top}Q_{e}(\hat{\vartheta}_k - g(q_k))$, is not convex in terms of $q_k$. Therefore, in the current formulation in \eqref{kin_opt_original}, it is significantly challenging to solve directly via QP due to the task tracking error term in the cost function.

To reformulate the above MPC problem as a QP problem, we linearize the nonlinear term along the nominal trajectory. By only considering the first term in the Taylor expansion, we approximate the nonlinear term as follows:
\begin{equation} \label{approx_tracking}
    \hat{\vartheta}_{k} - g(q_k) \approx \hat{J}_{k}( \hat{q}_{k} - q_k),
\end{equation}
where $\hat{J}_{k} = J(\hat{q}_{k})$ and $\hat{q}$ represents a nominal joint trajectory obtained by solving the inverse kinematics problem. We consider $\hat{J}_{k}$ as a stack of projected Jacobian matrices for $n_t$ tasks such as $J_{k}^{(j|\textrm{prev})}$ where $j\in \{1, \cdots, n_t\}$ in \eqref{cmd_kin}. 

In turn, we convexify the cost function by using the above approximation and the linear models for the joint velocity and acceleration:
\begin{align}
    \mathcal{J} &\approx \left(\sum_{k=i}^{i+n_p} q_{k}^{\top} \hat{J}_{k}^{\top}Q_{e} \hat{J}_{k}q_{k} - 2 \hat{q}_{k}^{\top}\hat{J}_{k} Q_{e} q_{k}\right) + \mathbf{q}^{\top} \mathbf{S}_v \mathbf{Q}_d \mathbf{S}_v \mathbf{q} \nonumber \\
    &\quad+ 2 \mathbf{v}^{\top} \mathbf{Q}_d \mathbf{S}_v \mathbf{q} + \mathbf{q}^{\top} \mathbf{S}_a \mathbf{Q}_a \mathbf{S}_a \mathbf{q} + 2 \mathbf{a}^{\top} \mathbf{Q}_a \mathbf{S}_a \mathbf{q} \\
    &= \mathbf{q}^{\top} \mathbf{Q} \mathbf{q} + 2 \mathbf{p}^{\top} \mathbf{q}, \nonumber 
\end{align}
where $\mathbf{Q}$ and $\mathbf{p}$ are the appropriate matrix and vector with $\mathbf{Q}_{d} = \textrm{diag}(Q_d, \cdots, Q_d)$ and $\mathbf{Q}_{a} = \textrm{diag}(Q_a, \cdots, Q_a)$. Now the cost function is expressed in a quadratic form. Using the above cost function, we reformulate the MPC as follows:
\begin{equation} \label{kin_mpc}
\begin{split}
    \min_{\mathbf{q}} \quad& \mathbf{q}^{\top} \mathbf{Q} \mathbf{q} + 2 \mathbf{p}^{\top} \mathbf{q} \\
    \textrm{s.t.} \quad& \dot{\mathbf{q}}_{\min} \leq \mathbf{S}_{v} \mathbf{q} + \mathbf{v} \leq \dot{\mathbf{q}}_{\max}, \\
    & \mathbf{q}_{\min} \leq \mathbf{q} \leq \mathbf{q}_{\max},
\end{split}    
\end{equation}
where $(.)_{\min}$ and $(.)_{\max}$ denote the minimum and maximum values for the bound of $(.)$. The optimization problem in \eqref{kin_mpc} is solvable via QP, allowing us to rapidly compute the joint position command. We will analyze the detailed computation time in terms of the prediction horizons in Section IV.

To verify stability of the formulated MPC, we enforce additional constraints for the terminal joint position and velocity. If the final prediction of \eqref{kin_opt_original} includes the terminal time step $(i+n_p = N)$, we need to insert additional inequality constraints $\hat{q}_N - \epsilon_{q} \leq q_{N} \leq \hat{q}_N - \epsilon_{q}$ and $\hat{q}_N - \epsilon_{v} \leq q_{N} \leq \hat{q}_N - \epsilon_{v}$ where $\epsilon_{q}$ and $\epsilon_{v}$ are allowable small boundaries for the stability verification. These constraints guarantee that the system controlled by the MPC is stable, assuming that the nominal state at the terminal time step is quasi-static or at equilibrium.  

\subsection{Nominal Trajectory via Inverse Kinematics}
The linearization of $g(q)$ relies on a nominal trajectory, which can be obtained through inverse kinematics. One simple method to solve the inverse kinematics problem is to employ the pseudoinverse of the Jacobian \cite{lee2012intermediate}. In the $k$-th time step, the joint velocity command for executing $n_t$ tasks is computed as
\begin{equation} \label{cmd_kin}
    \begin{split}
        \dot{q}_{k}^{(j)} = \dot{q}_{k}^{(j-1)} + \left( J_{k}^{(j|\textrm{pre})}\right)^{\dag}\left( k^{(j)} e_{k}^{(j)} - J_{k}^{(j)} \dot{q}_{k}^{(j-1)}\right),
    \end{split}
\end{equation}
where $\dot{q}_{k}^{(0)} = 0$, $J_{k}^{(j|\textrm{pre})} = J_{k}^{(j)}N_{k}^{(j-1)}$ and $N_{k}^{(j)} = N_{k}^{(j-1)} - \left(J_{k}^{(j|\textrm{pre})}\right)^{\dag} J_{k}^{(j|\textrm{pre})}$. In addition, the superscript $(j)$ denotes the properties for the $j$-th task. For instance, $e^{(j)}$ and $k^{(j) >0}$ are the task error and constant gain for the $j$-th task, respectively. By recursively computing the above equation \eqref{cmd_kin}, the command velocity for the hierarchical tasks is $\dot{q}_{k}^{(\textrm{cmd})} = \dot{q}_{k}^{(n_t)}$. When computing the pseudoinverse of $J$, we incorporate the compact Singular Value Decomposition (SVD) to prevent the system from diverging as follows:
\begin{equation}
    \begin{split}
        J = U\Sigma V^{\top} \approx U_r \Sigma_r V_r^{\top},
    \end{split}
\end{equation}
where $\Sigma_r$ represents the reduced singular value matrix by removing the smaller singular values than the pre-defined threshold. $U_r$ and $V_r$ are corresponding to the reduced $\Sigma_{r}$. Using the compact SVD, we compute the pseudoinverse of the Jacobian as
\begin{equation}
    J^{\dag} = V_r \Sigma_{r}^{-1} U_r^{\top}.
\end{equation}
Using the above pseudoinverse of the Jacobian, we prevent the robotic system from becoming unstable near or at singular configurations. However, the joint position may abruptly change due to the heuristic threshold for the compact SVD. Our MPC approach is able to resolve this discontinuity issue via additive damping and acceleration terms in the cost function. In Section IV, we will compare the results of the simple inverse kinematics method with our MPC approach.  

\section{Dynamic MPC with Constraints}
\subsection{Nonlinear and Continuous-time Dynamic Model}
The rigid body dynamics equation of a manipulator is expressed as follows:
\begin{equation}
    M(q) \ddot{q} + b(\dot{q}, q) = u,
\end{equation}
where $u \in \mathbb{R}^{n}$, $M(q) \in \mathbb{S}_{++}^{n}$, and $b(\dot{q},q) \in \mathbb{R}^{n}$ denote the torque input, the mass/inertia matrix, and sum of Coriolis/centrifugal and gravitational forces, respectively. The forward and inverse dynamic equations are represented as
\begin{equation}
    \begin{split}
    \textrm{FD}(q, \dot{q}, u) =& M(q)^{-1} ( u - b(\dot{q}, q)) =\ddot{q}, \\
    \textrm{ID}(q, \dot{q}, \ddot{q}) =&  M(q)\ddot{q} + b(\dot{q},q) = u.
    \end{split}
\end{equation}
We use simplified notations $M$, $b$, $J$, and $\dot{J}$ for $M(q)$, $b(\dot{q},q)$, $J_{k}(q)$, and $\dot{J}_{k}(\dot{q},q)$. Now, defining a state $x = [ q^{\top}, \dot{q}^{\top}]^{\top} \in \mathbb{R}^{n_x}$ a continuous-time state space model can be expressed as follows:
\begin{equation}
\begin{split}
    \dot{x} =& f(x, u) = \left[\begin{array}{c} \dot{q} \\ \textrm{FD}(q, \dot{q}, u)\end{array}\right],
\end{split}
\end{equation}
where $f: \mathbb{R}^{n_x}\times \mathcal{U} \mapsto \mathbb{R}^{n_x}$ and is nonlinear. Since we want to formulate a QP problem, we need to linearize and discretize the above state-space model to formulate the MPC in the shape of the QP problem.

\subsection{Discrete and Linear State-Space Model}
We consider the finite-time horizon $\mathbf{T} = [t_i, t_f]$ and normalize the time domain by using a dilation coefficient $\sigma = t_i - t_f$. The normalized variable is defined as $\tau = \sigma^{-1}(t-t_i) \in [0,1]$ for the unit interval. Thus we have
\begin{equation}
    \dot{x}_{\tau}  = \frac{d x_{\tau}}{dt} = \frac{1}{\sigma}\frac{dx_{\tau}}{d\tau} = f( x_{\tau}, u_{\tau}).
\end{equation}
The dynamics model in the normalized time domain is linearized along a given nominal trajectory $(\hat{x}_{\tau}, \hat{u}_{\tau})$:
\begin{equation}
    d x_{\tau} \approx (A_{\tau} x_{\tau}  + B_{\tau} u_{\tau} + r_{\tau})d \tau,
\end{equation}
where $A_{\tau} = \sigma \left. \nabla_{x}f(x, u) \right|_{(\hat{x}_{\tau}, \hat{u}_{\tau})}$, $ B_{\tau} = \sigma \left. \nabla_{u}f(x, u) \right|$ $_{(\hat{x}_{\tau}, \hat{u}_{\tau})}$, and $r_{\tau} = \sigma  f(\hat{x}_{\tau}, \hat{u}_{\tau})-A_{\tau}\hat{x}_{\tau} - B_{\tau}\hat{u}_{\tau}$. In the above formulations of $A_{\tau}$ and $B_{\tau}$, we utilize the partial derivative of Lagrangian expressions of forward dynamics which are 
\begin{equation}
\begin{split}
    \frac{\partial \textrm{FD}}{\partial q} =& \frac{\partial M^{-1}}{\partial q} (u - b) - M^{-1}  \frac{\partial b}{\partial q},  \\
    \frac{\partial \textrm{FD}}{\partial \dot{q}} =& - M^{-1} \frac{\partial b}{\partial \dot{q}}, \quad  \frac{\partial \textrm{FD}}{\partial u} = M^{-1},
\end{split}
\end{equation}
and $\frac{\partial M^{-1}}{\partial q} = - M^{-1} \frac{\partial M^{-1}}{\partial q}M^{-1}$. From \cite{carpentier2018analytical}, the relationship between the derivatives of inverse and forward dynamics are 
\begin{equation}
    \begin{split}
        \left. \frac{\partial \textrm{FD}}{\partial \xi} \right|_{(\hat{q}_\tau, \hat{\dot{q}}_{\tau}, \hat{u}_{\tau})} = -M(q_{\tau}^{d})^{-1} \left. \frac{\partial \textrm{ID}}{\partial \xi} \right|_{(\hat{q}_\tau, \hat{\dot{q}}_{\tau}, \hat{\ddot{q}}_{\tau})},
    \end{split}
\end{equation}
where $\xi \in\{q, \dot{q}\}$. $\frac{\partial \textrm{ID}}{\partial q}$ and $\frac{\partial \textrm{ID}}{\partial \dot{q}}$ are directly obtained by the recursive Newton-Euler algorithm.\footnote{Implementation of these rigid-body dynamics and partial derivative computations is available in the open-source Pinocchio library:\\ \url{https://github.com/stack-of-tasks/pinocchio}} In this study, we employ the computation algorithm proposed in \cite{carpentier2018analytical} to obtain the partial derivative terms. 

We convert the continuous-time state space model to discrete time by integrating the above differential equation:
\begin{equation}
    \int_{\tau_k}^{\tau_k + \Delta \tau} dx_{\tau} =\int_{\tau_k}^{\tau_k + \Delta \tau} (A_{\tau} x_{\tau} + B_{\tau} u_{\tau} + r_{\tau}) d{\tau},
\end{equation}
where we set $\Delta \tau = \Delta t$. Then, the discrete-time state space model is obtained as follows:
\begin{equation}
    x_{k+1} = A_{k} x_{k} + B_{k} u_{k} + r_{k},
\end{equation}
where $A_{k} = A_{\tau_k}\Delta t + I$ , $B_{k} = B_{\tau_k}\Delta t$, and $r_{k} = r_{\tau_k}\Delta t$ with $k\in \{i, \cdots, i+n_p\}$. Considering the concatenated state and control input vectors: $\bm{x}_{i} = [x_{i}^{\top}, \: \cdots, \: x_{i+n_p}^{\top} ]^{\top}$, $\bm{u}_{i} = [u_{i}^{\top}, \: \cdots, \: u_{i+n_p-1}^{\top} ]^{\top}$, and $ \bm{r}_{i} = [r_{i}^{\top}, \: \cdots, \: r_{i+n_p-1}^{\top} ]^{\top}$, we formulate the discrete-time state model in a similar form to \cite{lee2020mpc} as
\begin{equation} \label{lin_state}
    \bm{x}_{i} = \mathbf{A}_{i}x_{i} + \mathbf{B}_{i}\bm{u}_{i} + \mathbf{D}_{i}\bm{r}_{i} ,
\end{equation}
where $\mathbf{A}_{i} = \Omega(i)$, 
\begin{equation}
    \begin{split}
        \mathbf{B}_{i} =& [\Omega(i+1)B_{i}, \: \cdots, \: \Omega(i+n_p)B_{i+n_p-1}],\\
        \mathbf{D}_{i} =& [\Omega(i+1),\: \cdots,\: \Omega(i+n_p)],\\
        \Omega(s) =& [ \Phi(i,s)^{\top},\: \cdots, \: \Phi(i+n_p,s)^{\top}]^{\top}. 
    \end{split}
\end{equation}
 In addition, the matrix $\Phi(j,s)$ is computed as follows:
 \begin{equation}
     \Phi(j,s) = \left\{\begin{array}{ll} A_{j-1}\cdots A_{s} & \textrm{when } j \geq s+1 \\
     I & \textrm{when } j = s \\
     0 & \textrm{otherwise.} \end{array} \right. 
 \end{equation}
The above linear state-space model in \eqref{lin_state} is rearranged in terms of a decision variable $\mathbf{z}_{i} = [\mathbf{x}_{i}^{\top},\: \mathbf{u}_{i}^{\top} ]^{\top} $ as
\begin{equation}
    \left[\begin{array}{cc} I & - \mathbf{B}_{i} \end{array} \right] \mathbf{z}_{i} = \mathbf{A}_{i}x_{i} + \mathbf{D}_{i} \mathbf{r}_{i}.
\end{equation}
Now, we consider the constraint corresponding to the nonlinear dynamics of robots as a linear constraint in terms of our decision variable $\mathbf{z}_i$.

\begin{figure*}
\centering
\includegraphics[width=\linewidth]{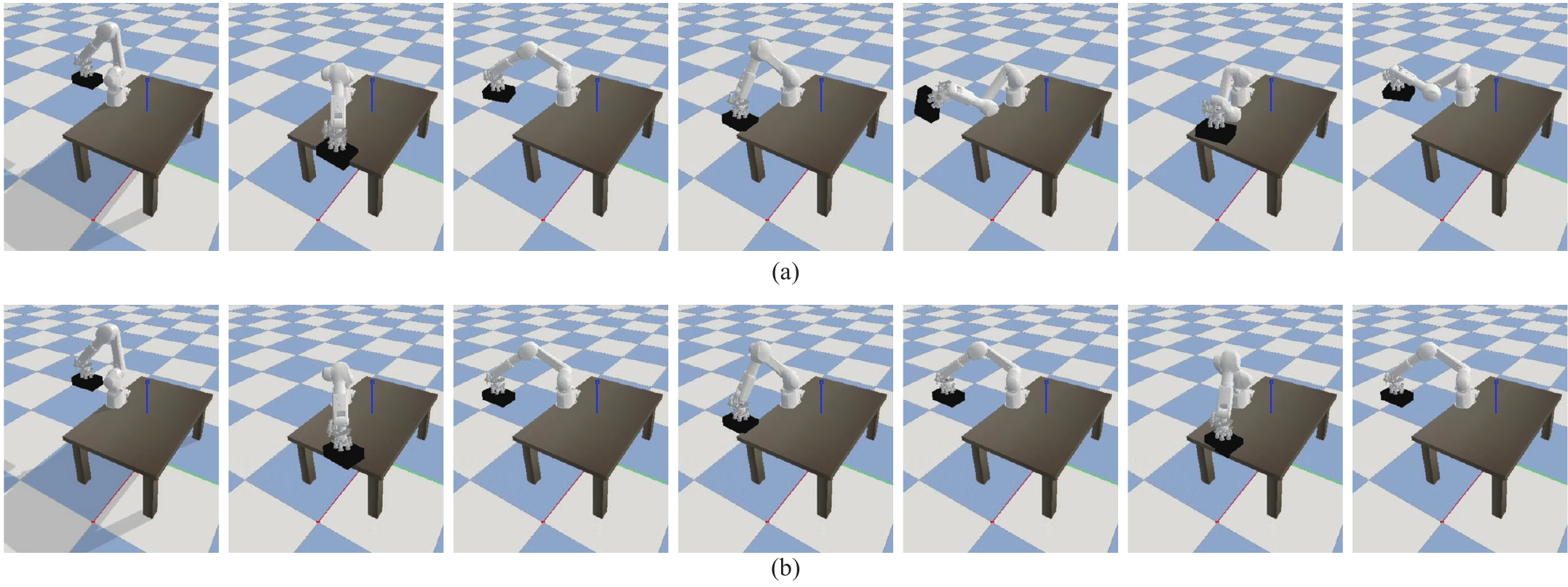}
\caption{{\bf Simulation snapshots}: (a) Task trajectory tracking using OSC, (b) Task trajectory tracking using the proposed MPC}
\label{Fig2}
\vspace{-10pt}
\end{figure*}

\subsection{MPC formulation}
Similar to the kinematic MPC formulation, we formulate dynamic MPC, including torque limit constraints. With prediction horizon $n_p$, the optimization problem is defined as
\begin{align}
    \min_{\mathbf{x}_{i}, \mathbf{u}_{i}} \quad & \mathcal{J} = \sum_{k=i}^{i+n_p} (\hat{\vartheta}_k - g(q_k))^{\top}Q_{e}(\hat{\vartheta}_k - g(q_k)) \nonumber \\
    & \qquad + \dot{q}_{k}^{\top}Q_{d} \dot{q}_{k} + u_{k}^{\top}Q_u u_{k}, \nonumber\\
    \textrm{s.t.}\quad& x_{k+1} = A_k x_k + B_k u_k +r_k ,\\
        & u_{\min} \leq u_{k} \leq u_{\max}, \quad \forall k \in \{ i, \cdots, i+n_p-1\}, \nonumber\\
    & \dot{q}_{\min} \leq \dot{q}_{j} \leq \dot{q}_{\max}, \nonumber\\
    & q_{\min} \leq q_{j} \leq q_{\max}, \quad \forall j \in \{ i, \cdots, i+n_p\}, \nonumber
\end{align}
where $Q_{u}$ denotes the weighting matrix for the torque input term. We approximate the tracking error term in the cost function by using \eqref{approx_tracking}: 
\begin{align}
    \mathcal{J} &\approx \left(\sum_{k=i}^{i+n_p} q_{k}^{\top} \hat{J}_{k}^{\top}Q_{e} \hat{J}_{k}q_{k} - 2 \hat{q}_{k}^{\top}\hat{J}_{k} Q_{e} q_{k}\right) + \dot{\mathbf{q}}_{i}^{\top} \mathbf{Q}_d \dot{\mathbf{q}}_{i} \nonumber \\
    & \quad + \mathbf{u}_{i}^{\top} \mathbf{Q}_u \mathbf{u}_{i} \\
    &= \mathbf{z}_{i}^{\top} \mathbf{W}_{i} \mathbf{z}_{i} + 2 \mathbf{w}_{i}^{\top} \mathbf{z}_{i}, \nonumber
\end{align}
where $\mathbf{Q}_{u} = \textrm{diag}(Q_u, \cdots, Q_u)$. In addition, $\mathbf{W}_{i}$ and $\mathbf{w}_{i}$ are the proper matrix and vector for the quadratic cost function in terms of the decision variable $\mathbf{z}_i$. Finally, we reformulate the optimization problem in terms of the decision variable $\mathbf{z}_i$ as follows:
\begin{equation}
    \begin{split}
        \min_{\mathbf{z}_i} \quad& \mathbf{z}_{i}^{\top} \mathbf{W}_{i} \mathbf{z}_{i} + 2\mathbf{w}_{i}^{\top}\mathbf{z}_{i} \\
        \textrm{s.t.}\quad&     \left[\begin{array}{cc} I & - \mathbf{B}_{i} \end{array} \right] \mathbf{z}_{i} = A_{i}x_{i} + \mathbf{D}_{i} \mathbf{r}_{i} ,\\
        & \mathbf{z}_{\min} \leq \mathbf{z}_i \leq \mathbf{z}_{\max},
    \end{split}
\end{equation}
where $\mathbf{z}_{\min}$ and $\mathbf{z}_{\max}$ are the minimum and maximum bounds for the decision variable $\mathbf{z}_{i}$. Although the dimension of the decision variable in the dynamic MPC ($\dim(\mathbf{z}_i) = 3 n_p n - n$) is larger than that of kinematic MPC($\dim(\mathbf{q}) = n_p n$), the reformulated optimization problem is solvable via QP, which is faster than nonlinear MPC. Similar to the kinematic MPC, we enforce the additional state constraints for the stability verification if the last prediction step is at the end of the trajectory.

\subsection{Nominal Trajectory via Inverse Dynamics}
The proposed MPC relies on the linearized state-space model of robot dynamics. For this reason, it is important to generate a realistic nominal trajectory, which can be done using an inverse dynamics controller, for example employing OSC \cite{sentis2005synthesis}. When computing a dynamically consistent inverse of the Jacobian, we also utilize the compact SVD to prevent the robot from becoming unstable near singular configurations.   

\section{Simulations and Experiments}
In this section, we validate the proposed QP-based MPC approach using two Kawasaki manipulators: RS007N (simulation) and RS020N (experiment). In the simulation, the robot is controlled by torque commands. On the other hand, the real robot is controlled by a joint position controller provided by Kawasaki control boxes. We validate the effectiveness of the devised dynamic MPC by controlling the manipulator (RS007N) with a heavy payload in the Pybullet simulation environment \cite{coumans2021}. We use \textit{QuadProg}++\footnote{QuadProg++: \url{https://github.com/liuq/QuadProgpp}}, which is based on Goldfarb-Idinani active-set dual method, to solve the formulated QP problems on a laptop with a i7-8650U CPU and 16 GB of RAM. In addition, the experimental work is demonstrated using the real robot (an RS020N with supporting software provided by Dexterity).\footnote{Dexterity, Inc.: \url{https://www.dexterity.ai}} 

\subsection{Handling a payload with torque limits}
In this simulation, we aim to track the desired end-effector's trajectory used in Dexterity applications while picking and placing parcels. The weight of the parcel is $12$ kg, which is heavier than those of normal packages. We consider the joint torque limits as $[239,\: 239,\: 124.5,\: 32,\: 40.96, \:25.6 ]$ $Nm$. In addition, it is assumed that the suction gripper's capacity is enough to handle the payload. For OSC, the end-effector position task is higher than the end-effector orientation task. We set the PD gains for the above tasks as $K_{\textrm{pos}}^{p} = [100,\: 100,\: 100 ]$, $K_{\textrm{pos}}^{d} = [7,\: 13,\:7 ]$, $K_{\textrm{ori}}^{p} = [20,\: 20,\: 20]$, and $K_{\textrm{ori}}^{d} =[1.5,\: 1.5,\: 1.5]$. In addition, the gains of the joint impedance controller are $K_{\textrm{imp}}^{p} =[100,\:100,\:100,\:50,\:50,\:1]$ and $K_{\textrm{imp}}^{d} =[3,\:5,\: 5,\: 0.2,\: 0.2,\: 0.1]$. We use two diagonal weighting matrices $Q_{e} = \textrm{diag}(10, \cdots, 10)$ and $Q_{d} = \textrm{diag}(0.0001, \cdots, 0.0001)$ without $Q_u$.

\begin{figure}
\centering
\includegraphics[width=\linewidth]{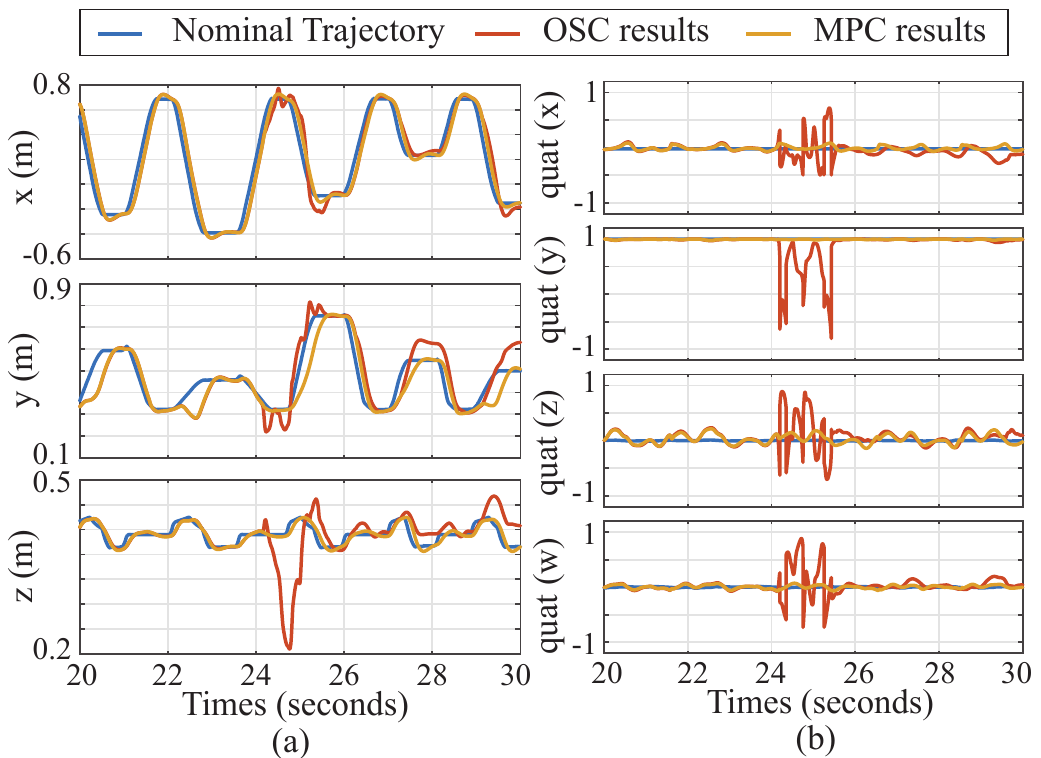}
\caption{{\bf $\textrm{SE}(3)$ of the end-effector}: (a) the position of the end-effector, (b) the orientation of the end-effector.}
\label{Fig3}
\vspace{-10pt}
\end{figure}

Fig. \ref{Fig2} (a) and (b) show the snapshots of the simulations demonstrating OSC and MPC with the same task trajectory, respectively. As shown in Fig. \ref{Fig2} (a), the robot controlled by OSC abruptly changes the configuration due to the heavy payload (see the $5$-th snapshot in Fig. \ref{Fig2} (a)). On the other hand, the proposed MPC prevents the rapid change of the configuration while tracking the given task trajectory. The above behavioral difference is shown in Fig. \ref{Fig3}, including the desired task trajectory (blue line) and the results controlled by OSC (orange line) and MPC (yellow line). More specifically, the end-effector's position in the z direction and its orientation are significantly fluctuated between $24$ seconds to $26$ seconds. Although the MPC results are not perfectly tracking the trajectory due to the payload, the overall tracking error of MPC is much smaller than OSC. 

\begin{figure}
\centering
\includegraphics[width=\linewidth]{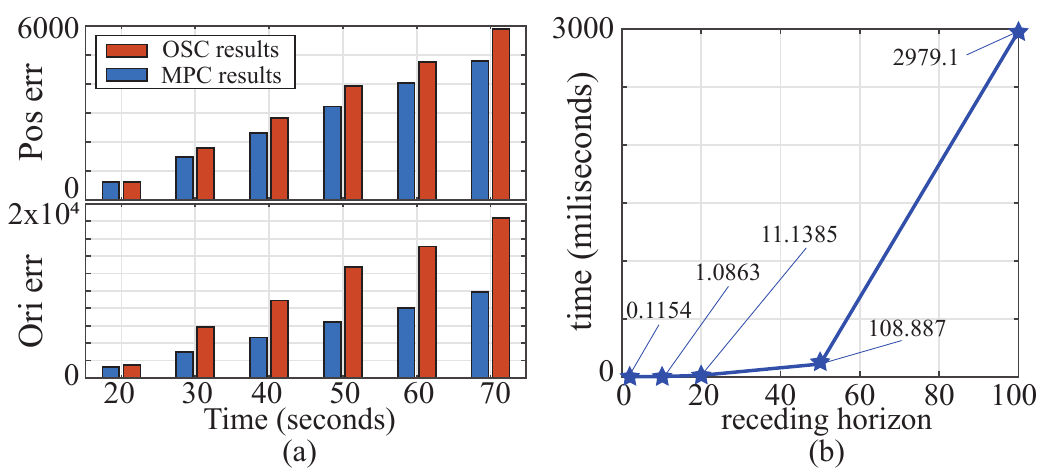}
\caption{(a) accumulated errors of the tasks, (b) the computation time in terms of the receding horizon}
\label{Fig4}
\end{figure}

The detailed comparison of the tracking error is based on the accumulated tracking error norm defined as $\text{err} = \sum_{k=0}^{n_{a}} \| e_{k}^{(\textrm{pos, ori})}\|_2$ where $n_a$ denotes the time horizon for accumulating the norm of task errors. The accumulated errors of two main tasks are described in Fig \ref{Fig4} (a). Overall, the tracking errors of both tasks controlled by the proposed MPC are clearly smaller than those controlled by OSC for all time. As shown in Fig. \ref{Fig4} (a), the gap between the orientation errors increases significantly since the orientation task is lower prioritized in the hierarchical control. In addition, we analyze the computation time in terms of the receding horizon as presented in Fig. \ref{Fig4} (b). The proposed MPC approach is formulated and solved as a QP, so the computation time depends on the dimension of the decision variables. The average computation time exponentially increases with increased receding horizon. At least $10$ receding horizon ($10$ milliseconds) can be implemented with $1$ kHz closed-loop controller, even given the limited specs of the laptop.

\begin{figure}
\centering
\includegraphics[width=\linewidth]{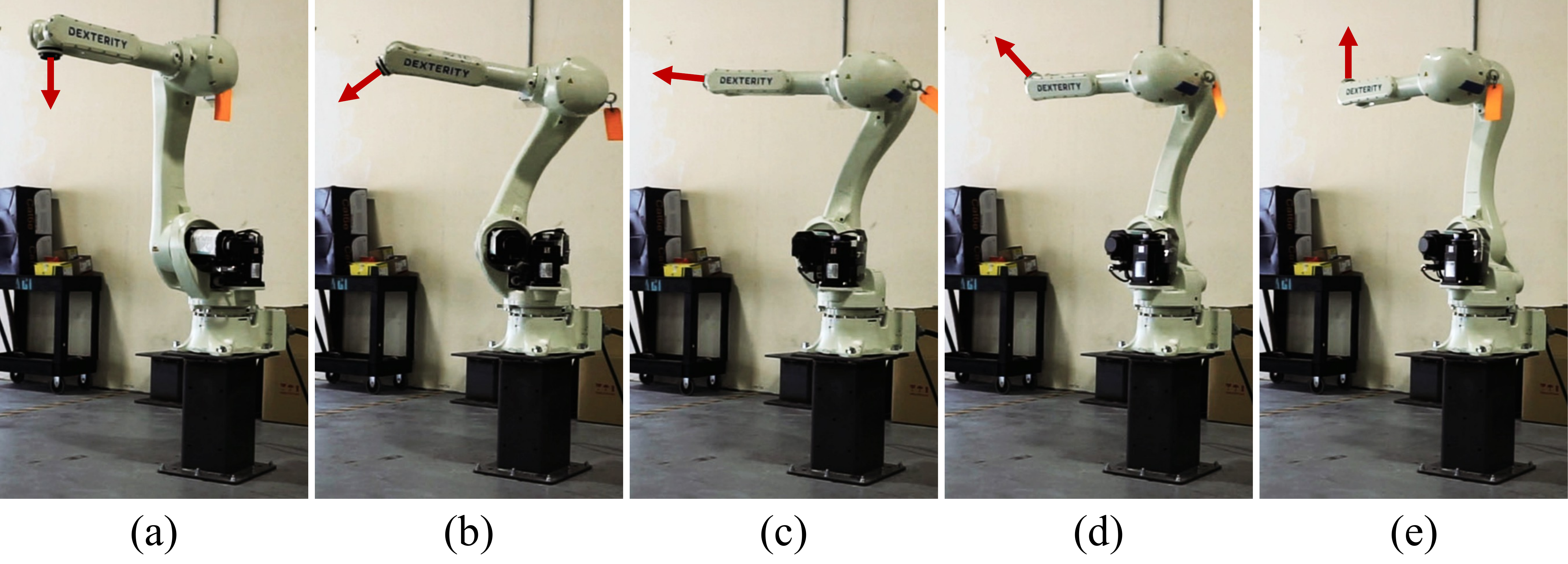}
\caption{{\bf Experiment snapshots}: The red arrows visualize the orientation of the end-effector. (a) initial configuration, (b) moving toward singular configuration, (c) avoiding singularity, (d) moving after avoiding singularity, (e) final configuration}
\label{Fig5}
\vspace{-10pt}
\end{figure}

\subsection{Singularity-free manipulation}
We verify our MPC by demonstrating a real experiment using a RS020N robot. The initial and final configurations are $[0,\: 0,\: -\frac{\pi}{2},\: 0,\: -\frac{\pi}{2}, \: \frac{\pi}{2}]$ rad and $[\frac{\pi}{2},\: 0,\: -\frac{\pi}{2} ,\: 0,\: \frac{\pi}{4},\: \frac{\pi}{2}]$ rad, respectively. The end-effector's position and orientation trajectories are generated using cubic spline and quaternion interpolation with $4$-second time durations. To perform the task trajectories, the robot must pass through a singularity when joint 5 is at zero. The weighting matrices for MPC are $Q_e = \textrm{diag}(2000,\cdots, 2000)$ and $Q_d = \textrm{diag}(0.01, \cdots, 0.01)$ without $Q_u$. Since the embedded joint space controller controls the real robot, we execute the dynamic simulation using the torque command and then generate the joint trajectory based on the measured joint position from the simulation. 

\begin{figure}
\centering
\includegraphics[width=\linewidth]{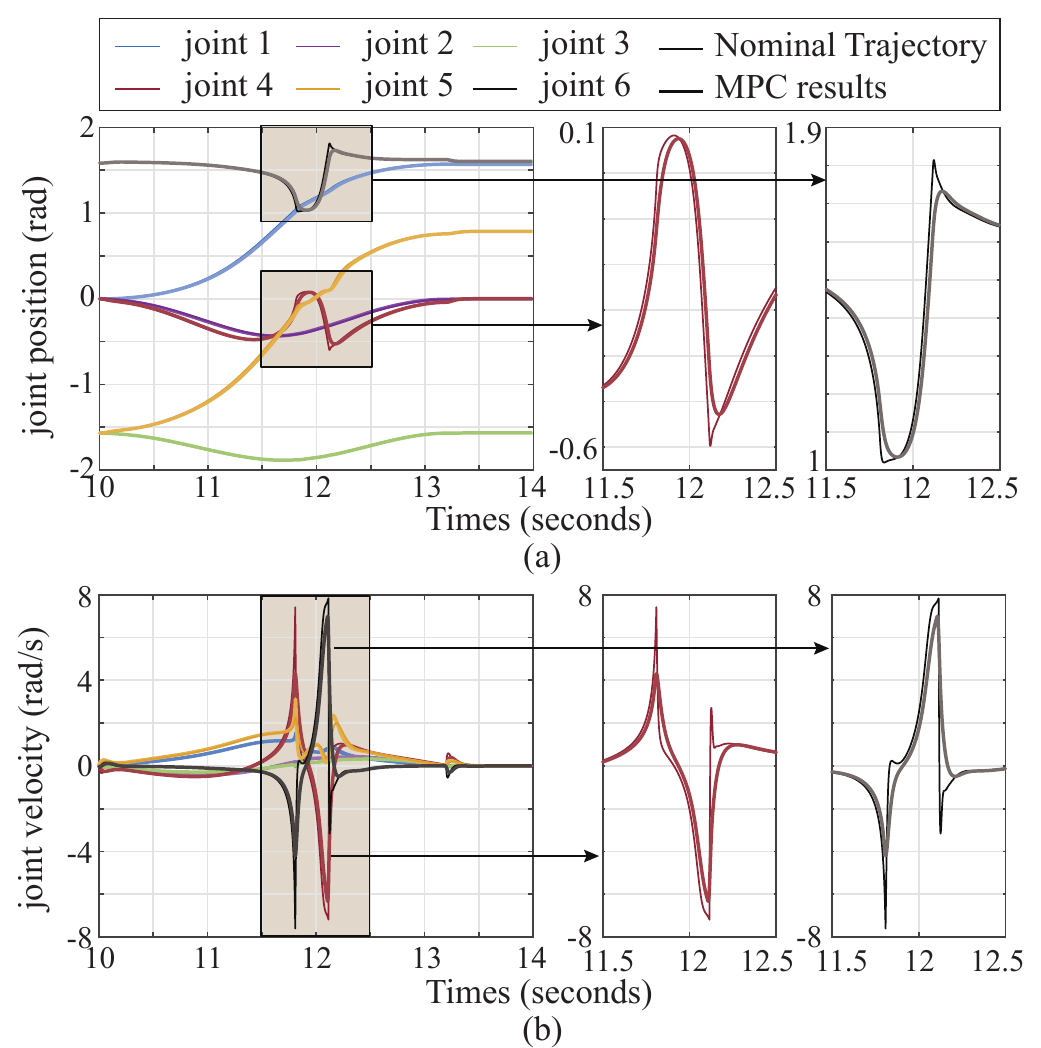}
\caption{{\bf Experiment results}: (a) joint position, (b) joint velocity. The plots around the singularity are magnified in the right-sided plots.}
\label{Fig6}
\vspace{-10pt}
\end{figure}

Fig. \ref{Fig5} represents the snapshots of the experiment using the proposed MPC. While moving from the initial configuration (Fig. \ref{Fig5} (a)) to the final one (Fig. \ref{Fig5} (e)), the robot avoids the singularity by twisting the last three joints (Fig. \ref{Fig5} (c)). On the other hand, the robot terminates the operation near the singularity when using a classical OSC controller. The measured joint position and velocity are shown in Fig. \ref{Fig6}. The regular and bold lines in Fig. \ref{Fig6} are the nominal trajectories computed by OSC and the results controlled by the proposed MPC, respectively. The proposed MPC generates much more smooth and stable behavior near the singular configuration than the conventional OSC with the compact SVD. In particular, we observe significant smoothness in the velocity level compared with the nominal trajectory as shown in Fig \ref{Fig6} (b). These experimental results show that the proposed MPC effectively avoids singularity while executing multiple tasks.

\section{Conclusion}
This paper proposes a real-time MPC framework to execute hierarchical tasks with low-level feedback controllers considering kinematics and dynamics and moving safely through singularity. Our proposed approach consists of first generating nominal trajectories using hierarchical control, linearizing along the nominal trajectories, and optimizing via a QP-based MPC formulation. We analyze the computation time of our framework for the underlying receding horizon and achieve real-time MPC rates of $1$ kHz for feedback control. The simulation and experiment results show that our proposed optimization framework successfully handles heavy payload tasks while fulfilling torque limits and efficiently avoids robot singularities to guarantee the achievement of smooth motions. 

In the future, we will employ the proposed MPC framework to more complicated and realistic scenarios for logistics. Also, we will consider learning the linearized model more accurately to reduce the tracking error with unknown and time-varying payloads.  





\renewcommand*{\bibfont}{\footnotesize}
\printbibliography 

\end{document}